\documentclass[11pt,a4paper]{article}
\usepackage[hyperref]{emnlp2020}
\usepackage{times}
\usepackage{latexsym}
\usepackage{xspace}

\usepackage{microtype}

\aclfinalcopy 
\def\aclpaperid{2800} 

\usepackage{balance}
\usepackage{amsfonts}
\usepackage{amsmath}
\usepackage{amssymb}
\usepackage{amsthm}

\usepackage{graphicx}
\usepackage[utf8]{inputenc} 
\usepackage[T1]{fontenc}    
\usepackage{hyperref}       
\usepackage{url}            
\usepackage{booktabs}       
\usepackage{nicefrac}       
\usepackage{microtype}      
\usepackage{lipsum}
\usepackage{bm}
\usepackage{subfigure}
\usepackage{hanging}
\usepackage{algorithm}
\usepackage{algpseudocode}
\usepackage{graphics}
\usepackage{epsfig}

\usepackage{nomencl}
\makenomenclature

\usepackage{etoolbox}

\usepackage{hyperref}
\usepackage{listings}
\newcommand{\sep}{\vspace{0.3ex}\noindent}
\usepackage{multirow}
\usepackage[normalem]{ulem}
\useunder{\uline}{\ul}{}

\usepackage{xargs}                      
\usepackage{cleveref}
\usepackage[colorinlistoftodos,prependcaption,textsize=tiny]{todonotes}

\crefname{section}{§}{§§}
\Crefname{section}{§}{§§}

\newcommandx{\yli}[2][1=]{\todo[linecolor=blue,backgroundcolor=blue!25,bordercolor=blue,#1]{#2}}
\newcommandx{\wendi}[2][1=]{\todo[linecolor=yellow,backgroundcolor=yellow!25,bordercolor=yellow,#1]{#2}}

\newcommand{\ie}{\emph{i.e.}\xspace} 
\newcommand{\eg}{\emph{e.g.}\xspace} 

\def \c {\mathbf{c}}
\def \d {\mathbf{d}}

\def \B {\mathbf{B}}
\def \D {\mathbf{D}}

\def \C {\mathcal{C}}

\def \R {\mathcal{R}}
\def \S {\mathcal{S}}

\theoremstyle{definition}
\newtheorem{definition}{Definition}
\newtheorem{example}{Example}

\title{Denoising Multi-Source Weak Supervision for Neural Text Classification}

\author{ Wendi Ren\textsuperscript{1}, Yinghao Li\textsuperscript{1}, Hanting Su\textsuperscript{2}, David Kartchner\textsuperscript{1}, Cassie Mitchell\textsuperscript{1} \and Chao Zhang\textsuperscript{1} \\
\textsuperscript{1} Georgia Institute of Technology, Atlanta, USA \\
\textsuperscript{2} Renmin University, Beijing, China \\
\texttt{ \{wren44, yinghaoli, david.kartchner, chaozhang\}@gatech.edu } \\
\texttt{ suhanting@ruc.edu.cn } \\
\texttt{cassie.mitchell@bme.gatech.edu} } 

\date{}

\begin{document}
\maketitle

\begin{abstract}

We study the problem of learning neural text classifiers without using any
labeled data, but only easy-to-provide rules as multiple weak
supervision sources. This problem is challenging because rule-induced weak
labels are often noisy and incomplete. To address these two challenges, we design a label denoiser, which estimates the source reliability using a conditional soft attention mechanism and then reduces label noise by aggregating rule-annotated weak labels. The denoised pseudo labels then supervise a neural classifier to predicts soft labels for unmatched samples, which address the rule coverage issue.
We evaluate our model on five benchmarks for sentiment, topic, and relation
classifications. The results show that our model outperforms state-of-the-art
weakly-supervised and semi-supervised methods consistently, and achieves
comparable performance with fully-supervised methods even without any labeled
data. Our code can be found at \url{https://github.com/weakrules/Denoise-multi-weak-sources}.
\end{abstract}

\section{Introduction}

Many NLP tasks can be formulated as text classification problems, such as sentiment analysis \citep{Badjatiya-2017-deep-learning}, topic classification \citep{zhang2015character}, relation extraction \citep{krebs-etal-2018-semeval} and question answering like slot filling \citep{pilehvar2018wic}.
Recent years have witnessed the rapid development of deep neural networks (DNNs)
for this problem, from convolutional neural network (CNN, \citealp{kim-2014-convolutional, kalchbrenner-etal-2014-convolutional}), recurrent neural network (RNN, \citealp{Lai-etal-2015-recurrent}) to extra-large pre-trained language models \citep{devlin2018bert, dai-etal-2019-transformer, liu2019roberta}.
DNNs' power comes from their capabilities of fitting complex functions based on large-scale training data.
However, in many scenarios, labeled data are limited, and manually annotating them at a large scale is prohibitively expensive.

Weakly-supervised learning is an attractive approach to address the data sparsity problem.
It labels massive data with cheap labeling sources such as heuristic rules or knowledge bases.
However, the major challenges of using weak supervision for text classification are two-fold:
1) the created labels are highly noisy and imprecise.
The \emph{label noise} issue arises because heuristic rules are often too simple to capture rich contexts and complex semantics for texts;
2) each source only covers a small portion of the data, leaving the labels incomplete.
Seed rules have \emph{limited coverage} because they are defined over the most frequent keywords but real-life text corpora often have
long-tail distributions, so the instances containing only long-tail keywords cannot be annotated.

Existing works \citep{ratner2017snorkel, meng2018weakly, zamani2018neural, Awasthi2020Learning} attempt to use weak supervision for deep text classification.
\citet{ratner2017snorkel} proposes a data programming method that uses labeling functions to automatically label data and then trains discriminative models with these labels.
However, data annotated in this way only cover instances directly matched by the rules, leading to limited model performance on unmatched data.
\citet{meng2018weakly} proposes a deep self-training method that uses weak supervision to learn an initial model and updates the model by its own confident predictions.
However, the self-training procedure can overfit the label noise and is prone to error propagation.
\citet{zamani2018neural} solves query performance prediction (QPP) by boosting
multiple weak supervision signals in an unsupervised way. However, they choose
the most informative labelers by an ad-hoc user-defined criterion, which may not generalize to all the domains. 
\citet{Awasthi2020Learning} assumes that human labelers are over-generalized to increase the coverage, and they learn restrictions on the rules to address learning wrongly generalized labels. However, their method requires the specific formulation process of rules to indicate which rules are generated by which samples, so that it cannot deal with other kinds of labeling sources like knowledge bases or third-party tools.  

We study the problem of using multiple weak supervision sources (\eg, domain
experts, pattern matching) to address the challenges in weakly-supervised text
classification. While each source is weak, multiple sources can
provide complementary information for each other.
There is thus potential to leverage
these multiple sources to infer the correct labels by estimating source
reliability in different feature regimes and then aggregating weak labels.
Moreover, since each source covers different instances, it is more promising to
leverage multiple sources to
bootstrap on unlabeled data and address the label coverage issue.

Motivated by the above, we propose a model with two reciprocal components.
The first is a \emph{label denoiser}  with the conditional soft attention mechanism \citep{bahdanau2014neural} (\cref{sect:rule}).
Conditioned on input text features and weak labels, it first learns reliability scores for labeling sources, emphasizing the annotators whose opinions are informative for the particular corpus.
It then denoises rule-based labels with these scores.
The other is a \emph{neural classifier} that learns the distributed feature representations for all samples (\cref{sect:neural}).
To leverage unmatched samples, it is supervised by both the denoised labels and its confident predictions on unmatched data.
These two components are integrated into an end-to-end co-training framework,
benefiting each other through cross-supervision losses, including the rule denoiser loss, the neural classifier loss, and the self-training loss(\cref{sect:loss}).

We evaluate our model on four classification tasks, including sentiment
analysis, topic classification, spam classification, and information extraction.
The results on five benchmarks show that: 1) the soft-attention module
effectively denoises the noisy training data induced from weak supervision
sources, achieving $84\%$ accuracy for denoising; and 2) the co-training design
improves prediction accuracy for unmatched samples, achieving at least $9\%$
accuracy increase on them. In terms of the overall
performance, our model consistently outperforms SOTA weakly supervised methods
\citep{ratner2017snorkel,meng2018weakly,zamani2018neural}, semi-supervised
method \citep{tarvainen2017mean}, and fine-tuning method
\cite{howard2018universal} by $5.46\%$ on average.

\section{Preliminaries}
\label{sec:org16666be}


\subsection{Problem Definition}
\label{sec:org22244b6}

In \emph{weakly supervised} text classification, we do not have access to clean labeled data. Instead, we assume external knowledge sources providing labeling rules as weak supervision signals.




\begin{definition}[Weak Supervision]
  A weak supervision source specifies a set of labeling rules $\R = \{r_1, r_2,
  \ldots, r_k\}$. Each rule $r_i$ declares a mapping $f \rightarrow C$, meaning
  any documents that satisfy the feature $f$ are labeled as $C$.
\end{definition}

We assume there are multiple weak supervision sources providing complementary information for each other. A concrete example is provided below.

\begin{figure}[t]
    \centering
    \includegraphics[width = 0.48\textwidth]{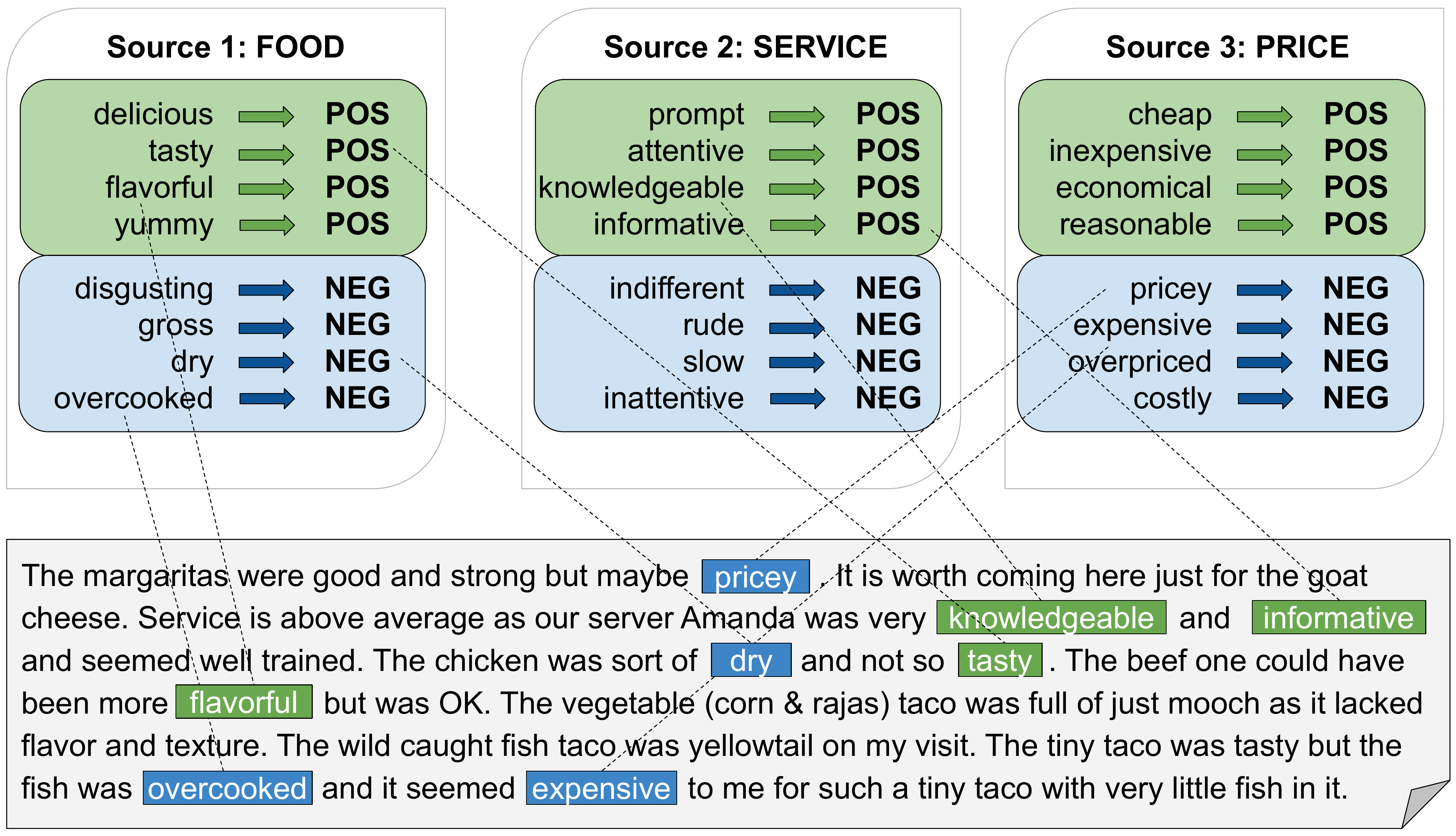}
    \caption{The annotation process for three weak supervision sources.
      ``POS'' and ``NEG'' are the labels for the sentiment analysis task.}
    \label{fig:LF_example}
\end{figure}

\begin{example}[Multi-Source Weak Supervision]
  Figure \ref{fig:LF_example} shows three weak sources for the sentiment
  analysis of Yelp reviews. The sources use `if-else' labeling functions to encode domain knowledge from different aspects. The samples that cannot be matched by any rules remain unlabeled.
\end{example}

\paragraph{Problem Formulation}
Formally, we have: 1) a corpus \(\D = \{\d_1, \ldots , \d_n\}\) of text documents; 2) a set \(\C = \{C_1, \ldots ,C_m\}\) of target classes; and 3) a set \(\S = \{\R_1, \R_2, \ldots, \R_k\}\) of weak annotators. Our goal is to learn a classifier from \(\D\) with only multiple weak supervision sources to accurately classify any newly arriving documents.
\subsection{Challenges}
\label{sec:orga76ee77}
Although the use of automatic weak annotators largely reduces human labeling efforts, using rule-induced labeled data has two drawbacks: \emph{label noise} and \emph{label incompleteness}.


Weak labels are noisy since user-provided rules are often simple and do not fully capture complex semantics of the human language.
In the Yelp example with eight weak supervision sources, the annotation accuracy is $68.3\%$ on average.
Label noise hurts the performance of text classifiers---especially deep classifiers---because such complex models easily overfit the noise.
Moreover, the source coverage ranges from $6.8\%$ to $22.2\%$.
Such limited coverage is because user-provided rules are specified over common lexical
features, but real-life data are long-tailed, leaving many samples unmatched by
any labeling rules.


\section{Our Method}
We begin with an overview of our method and then introduce its two key components as well as the model learning procedure.

\subsection{The Overall Framework}
\label{sect:overall}

\begin{figure}[tbp]
    \centering
    \includegraphics[width = 0.44\textwidth]{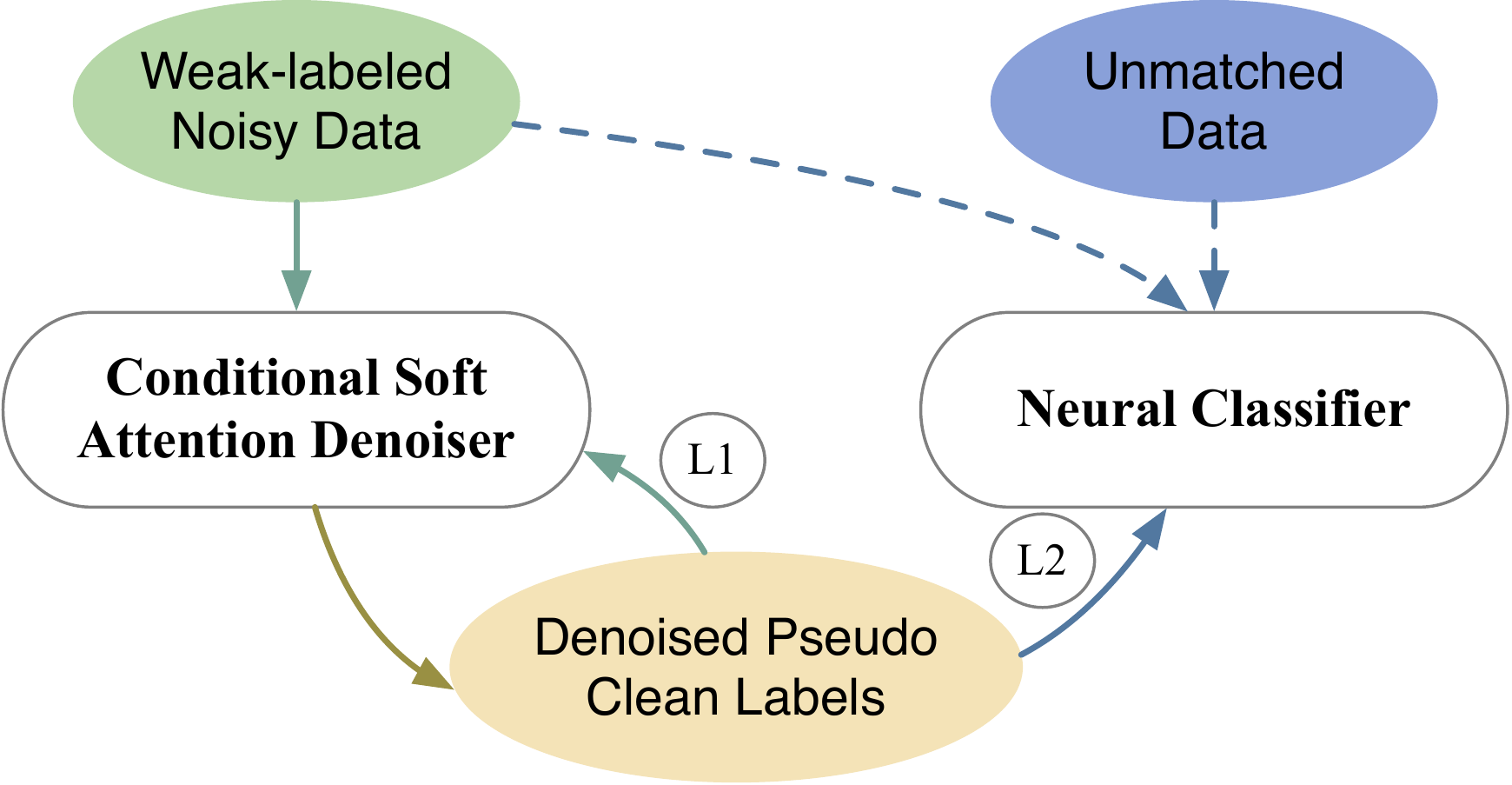}
    \caption{Overview of cross-training between the rule-based classifier and the neural classifier.}
    \label{fig:framework}
\end{figure}

Our method addresses the above challenges by integrating weak annotated labels from multiple sources and text data to an end-to-end framework with a label denoiser and a deep neural classifier, illustrated in Figure \ref{fig:framework}.

\paragraph{Label denoiser \& self-denoising}
We handle the label noise issue by building a label denoiser that iteratively denoises itself to improve the quality of weak labels.
This label denoiser estimates the source
reliability using a conditional soft attention mechanism, and then aggregates weak
labels via weighted voting of the labeling sources to achieve ``pseudo-clean'' labels. The reliability scores are conditioned on both rules and document feature representations. They effectively emphasize the opinions of informative sources while down-weighting those of unreliable sources, thus making rule-induced predictions more accurate.

\paragraph{Neural classifier \& self-training}
To address the low coverage issue, we build a neural classifier which learns distributed representations for text documents and classifies each of them, whether rule-matched or not. It is supervised by both the denoised weakly labeled data as well as its own high-confident predictions of unmatched data.


\begin{figure*}
    \centering
    \includegraphics[width = 1.0\textwidth]{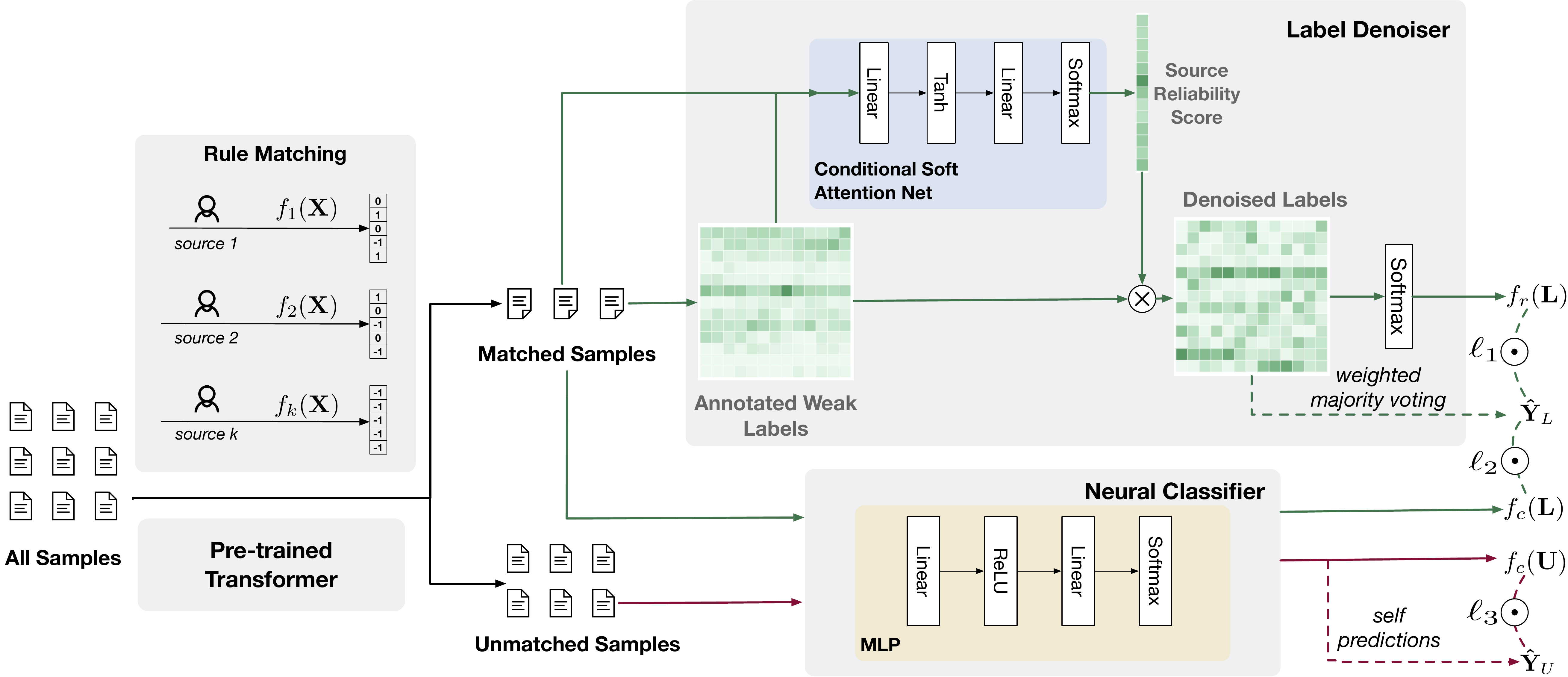}
    \caption{The detailed model architecture. Our model mainly consists of two
      parts: (1) the label denoiser, including the conditional soft attention reliability estimator and the instance-wise multiplication;
      (2) the neural classifier, which
      calculates sentence embedding using the pre-trained Transformer and makes
      classification.}
    \label{fig:model_overview}
\end{figure*}

\subsection{The Label Denoiser}
\label{sect:rule}
When aggregating multiple weak supervision sources, it is key for the model to
attend to more reliable sources, where source reliability should be conditioned
on input features. This will enable the model to aggregate multi-source weak
labels more effectively. Given $k$ labeling resources, we obtain the weak label
matrix \(\tilde Y \in \mathbb{R}^{n\times k}\) through rule matching. Specifically, as shown in the Rule Matching step of \ref{fig:model_overview}, by Definition 1, given one rule, if a document is matchable by that rule, it will be assigned with a rule-induced label C; otherwise, the document remains unlabeled, represented as -1. N rules thus generate N weak labels for each document. 
We then estimate the source reliability and aggregate complementary weak labels to obtain ``pseudo-clean'' labels.




\paragraph{Parameterization of source reliability}
We introduce a soft attention mechanism conditioned on both weak labels and feature representation, denoted as \(\bm B\), to estimate the source reliability.
Formally, we denote the denoised ``pseudo-clean'' labels by \( {\hat Y } = [\hat {y_{1}}, \ldots, \hat {y_{n}}]^T\) , and the initial ones ${\hat Y_0}$ are obtained by simple majority voting from \( {\tilde Y} \).

The core of the label denoiser is an attention net, a two-layer feed-forward neural network which predicts the attention score for matched samples. Formally, we specify a reliability score \(a_j\) for each labeling source to represent its annotation quality, and the score is normalized to satisfy $\sum_{j=1}^{k}a_{j} = 1$. For one document $\bm{d_i}$, its attention score \(q_{i,j}\) of one labeling source \(\R_j\) is:
\begin{equation}
\setlength{\abovedisplayskip}{0.12cm}
\setlength{\belowdisplayskip}{0.12cm}
  \begin{aligned}
  \hat{q}_{ij}=& W_2^{T} \tanh (W_{1} (\tilde y_{ij} + \bm B_i) ) ,
  \\ q_{ij}=& \frac{\exp (\hat{q}_{ij})}{\sum_{j} \exp (\hat{q}_{ij})} ,
  \end{aligned}
  \label{equ:reliability}
\end{equation}
where \(W_1, W_2\) denote the neural network weights and \(\rm tanh\) is the activation function. Thus, for each document, its conditional labeling source score vector $\bm A_i = [a_{i1}, a_{i2}, \ldots , a_{ik}]^T$ is calculated over matched annotators as
 $a_{ij}= q_{ij} \chi_C (\tilde y_{ij}>=0)$,
where $\chi_C$ is the indicator function. Then, we average the conditional source score $\bm A_i$ over all the $n$ matched samples to get the source reliability vector $\bm A$. The weight of \(j_{th}\) $(j = 1, 2,  \ldots, k)$ annotator is calculated as $a_{j}=\frac{1}{n} \sum_{i=1}^{n} a_{ij}$. Finally, We aggregate \(k\) reliability scores to get the reliability vector \(\bm A = [a_1, a_2, \ldots, a_k]^T\).

\paragraph{Denoising pseudo labels}
\label{sec:org199283f}
With the learned reliability vector \(\bm A\), we reweight the sources to get the weighted majority voted labels \({\hat Y}\) by ${\tilde Y_i} \otimes \bm A$. The denoised ``pseudo-clean'' label $\hat{y_i}$ is:
\begin{equation}
\setlength{\abovedisplayskip}{0.1cm}
\setlength{\belowdisplayskip}{0.1cm}
\begin{aligned}
\hat{y}_i=\arg \max _{C_r} \sum_{j=1}^{k} a_{j} \chi_{C}(\tilde y_{ij} == C_r), 
\\ \text{where } r = 1, 2, \ldots, m.
\end{aligned}
\label{equ:pseudo clean}
\end{equation}
The updated higher-quality labels \( {\hat Y} \) then supervise the rule-covered samples in \(\D_L\) to generate better soft predictions and guide the neural classifier later.

\paragraph{Rule-based classifier prediction}
At the epoch $t$ of our co-training framework, we learn the reliability score $\bm A(t)$ and soft predictions \(\bm{\hat Z}(t)\) supervised by ``pseudo-clean'' labels from the previous epoch \(  {\hat Y} (t-1) \). Then we renew ``clean-pseudo'' labels as \( {\hat Y} (t) \) using the score $\bm A(t)$ by \eqref{equ:pseudo clean}.

Specifically, given \(m\) target classes and \(k\) weak annotators, the prediction probability \(\bm{\hat z_i}\) for $\bm{d_i}$ is obtained by weighting the noisy labels \({\tilde Y_i}\) according to their corresponding conditional reliability scores $\bm A_i$:
$\bm{\hat z_i} = \text{softmax}({\tilde Y_i} \otimes \bm A_i)$,
where the masked matrix multiplication $\otimes$ (defined in \eqref{equ:prediction_rule}) is used to mask labeling sources that do not annotate document $i$, and we normalize the resultant masked scores via softmax:
\begin{equation}
\setlength{\abovedisplayskip}{0.12cm}
\setlength{\belowdisplayskip}{0.12cm}
    \begin{aligned}
  y_{ir} &= \sum_{j=1}^{k} a_{ij} \chi_{C}(\tilde y_{ij} == C_r) \\
 \hat{\mathbf{z}}_{ir} &= \frac{\exp(y_{ir})}{\sum_{r=1}^{m} \exp(y_{ir})}.
    \end{aligned}
    \label{equ:prediction_rule}
\end{equation}
We finally aggregate \(m\) soft adjusted scores to get the soft prediction vector \(\bm {\hat z_i} = [z_{i1}, \ldots, z_{im}]^T\).

\subsection{The Neural Classifier}
\label{sect:neural}
The neural classifier is designed to handle all the samples, including matched ones and unmatched ones. The unmatched corpus where the documents cannot be annotated by any source is denoted as \(\D_U\).
In our model, we use the pre-trained BERT~\cite{devlin2018bert} as our feature extractor, and then feed the text embeddings $\B$ into a feed-forward neural network to obtain the final predictions.
For \(\bm d_i\in {\D_L \cup \D_U}\), the prediction $\bm{\tilde {z}_{i}}$ is:
\begin{equation}
\setlength{\abovedisplayskip}{0.12cm}
\setlength{\belowdisplayskip}{0.12cm}
  \bm{\tilde {z}_{i}} = f_\theta(\bm B_{i} ; \theta_{w}),
  \label{equ:prediction_nn}
\end{equation}
where \(f_\theta\) denotes the two-layer feed-forward neural network, and \(\theta_w\) denotes its parameters.

\subsection{The Training Objective}
\label{sect:loss}

\noindent\textbf{The rule denoiser loss} $\ell_1$ is the loss of the rule-based classifier over \(\D_L\). We use the
``pseudo-clean'' labels  \(  {\hat Y} \) to self-train the label denoiser and define the loss $\ell_1$ as the negative log likelihood of \( {\hat y_{i}} \), 
\begin{equation}
\setlength{\abovedisplayskip}{0.12cm}
\setlength{\belowdisplayskip}{0.12cm}
  \ell_1 =-\sum_{i\in \D_L}  {\hat y_{i}}  \log \bm{\hat z_{i}}.
  \label{equ:L1}
\end{equation}

\noindent \textbf{The neural classifier loss} $\ell_2$ is the loss of the neural classifier over \(\D_L\).
Similarly, we regard the negative log-likelihood from the neural network outputs \(\bm {\tilde Z}\) to the pseudo-clean labels  \(  {\hat Y} \)  as training loss, formally 
\begin{equation}
\setlength{\abovedisplayskip}{0.12cm}
\setlength{\belowdisplayskip}{0.12cm}
  \ell_2=-\sum_{i\in \D_L} {\hat y_{i}}  \log \bm{\tilde z_{i}}.
  \label{equ:L2}
\end{equation}


\noindent \textbf{The unsupervised self-training loss} $\ell_3$ is the loss of the neural classifier over \(\D_U\).
To further enhance the label quality of \(\D_U\)
we apply the temporal ensembling strategy \cite{lainete2016}, which aggregates the predictions of multiple previous network evaluations into an ensemble prediction to alleviate noise propagation. For a document $\bm d_i\in \D_U$, the neural classifier outputs \(\bm{\tilde z_i}\) are accumulated into ensemble outputs \(\bm{Z_i}\) by updating \(\bm{{Z}_{i}} \leftarrow \alpha \bm{Z_{i}}+(1-\alpha) \bm{\tilde z_{i}}\), where \(\alpha\) is a term that controls how far the ensemble looks back into training history. We also need to construct target vectors by bias correction, namely \( \bm{p_i} \leftarrow \bm {Z_i} /(1-\alpha^{t})\), where $t$ is the current epoch. Then, we minimize the Euclidean distance between $\bm{p_i}$ and $\bm{\tilde z_i}$,
where 
\begin{equation}
\setlength{\abovedisplayskip}{0.12cm}
\setlength{\belowdisplayskip}{0.12cm}
  \ell_{3} = \sum_{i\in \D_U} \quad\| \bm{\tilde z_{i}} - \bm p_i \|^{2}
  \label{equ:L4}
\end{equation}

\paragraph{Overall Objective}
The final training objective is to minimize the overall loss $\ell$:
\begin{equation}
\setlength{\abovedisplayskip}{0.12cm}
\setlength{\belowdisplayskip}{0.12cm}
    \ell = c_1 \ell_{1} + c_2\ell_{2} + c_3\ell_3,
    \label{eq:loss_func}
\end{equation}
where $0 \le c_1 \le 1$, $0 \le c_2 \le 1$, and $0 \le c_3 \le 1$ are hyper-parameters for balancing the three losses and satisfy $c_1 + c_2 + c_3 = 1$.

\begin{algorithm}[tbp]
  \caption{Training process of our model}
  \label{alg::training}
  \begin{algorithmic}[1]
    \Require  $\D_L$, $\D_U$, $\C$, $\bm B$, ${\tilde Y}$, $g_w (x)$ and $f_\theta (x)$: feed-forward rule-based and nerual classifier with trainable parameters $W$ and $\theta$; $s$: number of training iteraions;
    \State ${\hat Y} \leftarrow {\hat Y_0}$, initialize by simple majority voting
    \For{$t \gets 1$ to $s$}
      \State $\bm A, \bm {\hat z_}{i \in \D_L} \leftarrow g_{w}( {\tilde
        Y_i}, \bm B_i, {\hat y_{i}})$ \Comment{learn reliability score and
        evaluate attention network output supervised by ``pseudo-clean'' labels
        from \eqref{equ:reliability} and \eqref{equ:prediction_rule}}
      \State ${\hat{y_i}} \leftarrow $\eqref{equ:pseudo clean} \Comment{renewed pseudo labels}
      \State $\bm {\tilde z_}{i \in {\D_L\cup \D_U}}  \leftarrow f_{\theta}(\bm {B_i}, {\hat{y_i}})$ \Comment{evaluate neural classifier output}
      \State update $\theta, W$ using ADAM by \eqref{eq:loss_func}
    \EndFor
    \State \textbf{return} $W, \theta$
  \end{algorithmic}
\end{algorithm}

\subsection{Model Learning and Inference}
\label{sect:learning}
Algorithm \ref{alg::training} sketches the training procedure. Two
classifiers provide supervision signals for both themselves and their peers,
iteratively improving their classification abilities. In the test phase, the corpus is sent into our model with the corresponding annotated noisy
labels. The final target $C_i$ for a document $i$ is predicted by ensembling the
soft predictions. If two predictions from the label denoiser and the neural classifier conflict with each other, we choose the one with higher confidence,
where the confidence scores are softmax outputs.



\begin{table}[tbp]
\begin{small}
	\begin{center}
		\begin{tabular}{ @{\hskip2.0pt} c @{\hskip2.0pt} |@{\hskip2.0pt} c @{\hskip2.0pt}|@{\hskip2.0pt} c @{\hskip2.0pt}|@{\hskip2.0pt} c @{\hskip2.0pt}|@{\hskip2.0pt} c@{\hskip2.0pt} |@{\hskip2.0pt} c@{\hskip2.0pt}| @{\hskip2.0pt} c@{\hskip2.0pt}| @{\hskip2.0pt} c@{\hskip2.0pt}}
			\toprule \bf Dataset & \bf Task & $C$ &\bf \#Train & \bf \#Dev & \bf \#Test & \bf Cover & \bf Acc. \\ \hline
			youtube &Spam & 2 &1k & 0.1k & 0.1k & 74.4 & 85.3\\ 
            imdb &Sentiment &2 &20k & 2.5k & 2.5k & 87.5 & 74.5 \\ 
            yelp &Sentiment& 2&30.4k & 3.8k & 3.8k & 82.8 & 71.5 \\ 
            agnews &Topic &4 &96k & 12k & 12k & 56.4 & 81.4 \\ 
            spouse & Relation & 2 & 1k & 0.1k & 0.1k &  85.9 & 46.5  \\
            \bottomrule
		\end{tabular}
	\end{center}
	\caption{Data Statistics. $C$ is the number of classes. Cover is fraction of rule-induced samples. Acc. refers to precision of labeling sources (number of correct samples / matched samples). Cover and Acc. are in \%.}
	\label{tab:statistics}
\end{small}
\end{table}

\section{Experiments}


\subsection{Experimental Setup}
\label{subsec:baselines}

\paragraph{Datasets and tasks} We evaluate our model on five widely-used text classification datasets, covering four different text classification tasks: \textbf{youtube}~\cite{alberto2015tubespam} (Spam Detection),
\textbf{imdb}~\cite{maas2011learning}, \textbf{yelp}~\cite{zhang2015character} (Sentiment Analysis),
\textbf{agnews}~\cite{zhang2015character} (Topic Classification), and \textbf{spouse}~\cite{ratner2017snorkel} (Relation Classification).
Table \ref{tab:statistics} shows the statistics of these datasets and the quality of weak labels (the details of each annotation rule are given in the \cref{sec:LFs}). Creating such rules required very light efforts, but is able to cover a considerable amount of data samples (\eg, 54k in agnews).

\paragraph{Baselines} We compare our model with the following advanced methods:
1) \textbf{Snorkel}~\cite{ratner2017snorkel} is a general weakly-supervised learning method that learns from multiple sources and denoise weak labels by a generative model;
2) \textbf{WeSTClass}~\cite{meng2018weakly} is
a weakly-supervised text classification model based on self-training;
3) \textbf{ImplyLoss}~\cite{Awasthi2020Learning} propose the rule-exemplar supervision and implication loss to denoise rules and rule-induced labels jointly;
4) \textbf{NeuralQPP}~\cite{zamani2018neural} is a boosting prediction framework which selects useful labelers from multiple weak supervision signals;
5) \textbf{MT}~\cite{tarvainen2017mean} is a semi-supervised model that uses Mean-Teacher method to average model weights and add a consistency regularization on the student and teacher model; and
6) \textbf{ULMFiT}~\cite{howard2018universal} is a strong deep text
classifier based on pre-training and fine-tuning.
7) \textbf{BERT-MLP} takes the pre-trained Transformer as the feature extractor and stacks a multi-layer perceptron on its feature encoder.

\subsection{Experimental Results}

\subsubsection{Comparison with Baselines}

We first compare our method with the baselines on five datasets. For fair
comparison, all the methods use a pre-trained BERT-based model for feature
extraction, and use the same neural architecture as the text classification
model. All the baselines use the same set of weak labels $\tilde Y$ for
model training, except for WeSTClass which only requires seed keywords as weak
supervision (we extract these keywords from the predicates of our rules).


\begin{table}[tbp]\small
    	\begin{center}
		\begin{tabular}
		{ @{\hskip2.0pt} c @{\hskip2.0pt} |@{\hskip2.0pt} c @{\hskip2.0pt}|@{\hskip2.0pt} c @{\hskip2.0pt}|@{\hskip2.0pt} c @{\hskip2.0pt}|@{\hskip2.0pt} c@{\hskip2.0pt} |@{\hskip2.0pt} c@{\hskip2.0pt}}
    \toprule
    \textbf{Method}  & \textbf{youtube} & \textbf{imdb}  & \textbf{ yelp }  & \textbf{agnews} & \textbf{spouse} \\
    \midrule
    Snorkel & 78.6 & 73.2 & 69.1 & 62.9 & 56.9 \\
    WeSTClass & 65.1 & 74.7 & 76.9 & 82.8 & 56.6 \\
    Implyloss & 93.6 & 51.1 & 76.3 & 68.5 & 68.3 \\
    NeuralQPP & 85.2 & 53.6 & 57.3 & 69.5 & 74.0 \\
    MT & 86.7 & 72.9 & 71.2 & 70.6 & 70.7\\
    ULMFiT & 56.1 & 70.5 & 67.3 & 66.8 & 72.4 \\
    BERT-MLP &77.0 & 72.5 &81.5 & 75.8 & 70.7 \\
    Ours &\textbf{94.9}&\textbf{82.9}& \textbf{87.5}&\textbf{85.7}& \textbf{81.3} \\
    \bottomrule
    \end{tabular}
    \end{center}
    \caption{Classification accuracy in the test set for all methods on five datasets.}
    \label{tab:main_result_ind}
\end{table}

Table \ref{tab:main_result_ind} shows the performance of all the methods on five
datasets. As shown, our model consistently outperforms all the baselines across
all the datasets. Such results show the strength and robustness of our model.
Our model is also very time-efficient ($4.5$ minutes on average) with trainable
parameters only from two simple MLP neural networks ($0.199$M trainable
parameters).

Similar to our methods,
Snorkel, NeuralQPP, and Implyloss also denoise the weak labels from
multiple sources by the following ideas: 1) Snorkel uses a generative modeling approach; 2) Implyloss
adds one regularization to estimate the rule over-generalizing issue, but it
requires the clean data to indicate which document corresponds to which rule. Without such information in our setting, this advanced baseline cannot perform
well; 3)  NeuralQPP selects the most informative weak labelers by boosting method.
The performance gaps verify the effectiveness of the our conditional soft attention design and co-training framework.

WeSTClass is similar to our method in that it also uses self-training to
bootstrap on unlabeled samples to improve its performance. The major advantage
of our model over WeSTClass is that it uses two different predictors (rule-based
and neural classifier) to regularize each other. Such a design not only better
reduces label noise but also makes the learned text classifier more robust.

Finally, ULMFiT and BERT-MLP are strong baselines based on language model fine-tuning.
MT is a well-known semi-supervised model which achieved inspiring results for image classification.
However, in the weakly supervised setting, they do not
perform well due to label noise.
The results show that ULMFiT and MT suffer from such label noise, whereas our model is noise-tolerant and more suitable in weakly supervised settings. Overall BERT-MLP performs the best and we further compare it with ours in more perspectives.

\subsubsection{Effectiveness of label denoising}
\begin{figure}[tbp]
    \centering
    \includegraphics[width = 0.48\textwidth]{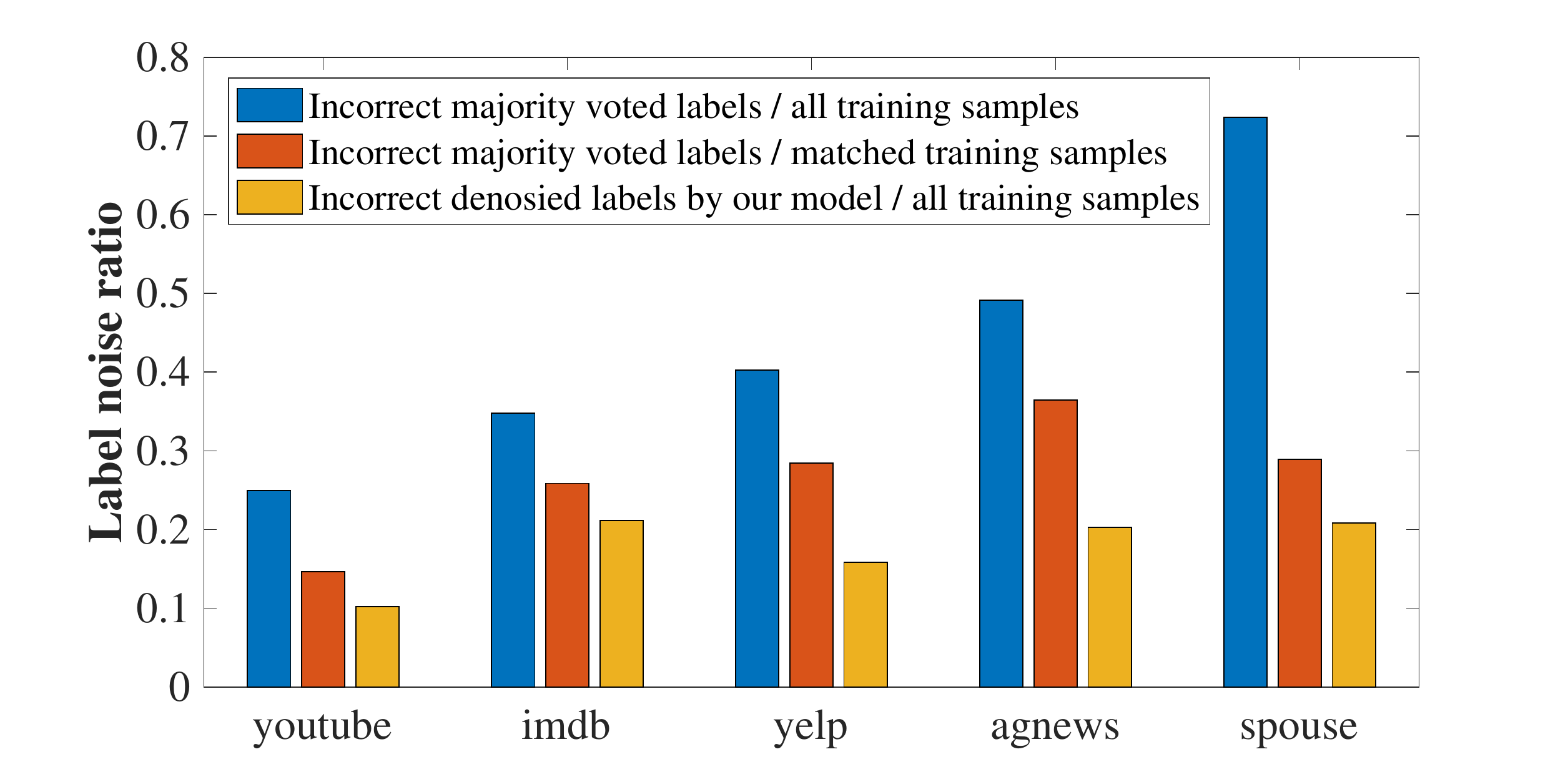}
    \caption{The label noise ratio of the initial majority voted labels and our denoised labels in the training set. }
    \label{fig:nosie_ratio}
\end{figure}

To study the effectiveness of label denoising, we first compare the label noise
ratio in training set given by the majority-voted pseudo labels ($\tilde Y$ defined in \cref{sect:rule}) and our denoised pseudo labels. Figure
\ref{fig:nosie_ratio} shows that after applying our denoising model, the label
noise is reduced by 4.49\% (youtube), 4.74\% (imdb), 12.6\% (yelp), 3.87\% (agnews) and 8.06\% (spouse) within the matched samples. If we count all the samples, the noise reduction is much more significant with 23.92\% on average.
Such inspiring results show the effectiveness of our model in
denoising weak labels.


\paragraph{Train a classifier with denoised labels}
We further study how the denoised labels benefit the training of supervised models. To this end, we feed the labels generated by majority voting and denoised ones generated by our model into two state-of-the-art supervised models: ULMFiT and BERT-MLP (described in \cref{subsec:baselines}).
Table~\ref{tab:noise_accuracy_sup} shows that denoised labels significantly improve the performance of supervised models on all the datasets.

\begin{table}[tbp]
\begin{small}
    	\begin{center}
		\begin{tabular}{ @{\hskip2.0pt} c @{\hskip2.0pt} |@{\hskip2.0pt} c @{\hskip2.0pt}|@{\hskip2.0pt} c @{\hskip2.0pt}|@{\hskip2.0pt} c @{\hskip2.0pt}|@{\hskip2.0pt} c@{\hskip2.0pt} |@{\hskip2.0pt} c@{\hskip2.0pt}|@{\hskip2.0pt} c@{\hskip2.0pt} }
    \toprule
    \textbf{Method}  & \textbf{Labels} & \textbf{youtube} & \textbf{imdb}  & \textbf{ yelp }  & \textbf{agnews} & \textbf{spouse} \\
    \midrule
    BERT+ & major & 77.0 & 72.5  & 81.5  & 75.8  & 70.7 \\
    MLP & ours & \textbf{89.8} & \textbf{80.2} & \textbf{85.8} & \textbf{84.3} & \textbf{78.0} \\\hline
    UlmFit & major & 56.1 & 70.5  & 67.3  & 66.8 & 72.4  \\
           & ours & \textbf{90.8}  & \textbf{81.6} & \textbf{85.9} & \textbf{84.7} & \textbf{81.3} \\
    \bottomrule
    \end{tabular}
    \end{center}
    \caption{Classification accuracy of two supervised methods with labels generated by majority voting and denoised ones generated by our model.}
    \label{tab:noise_accuracy_sup}
\end{small}
\end{table}


\subsubsection{Effectiveness of handling rule coverage}
\begin{figure}[tbp]
    \includegraphics[width = 0.48\textwidth]{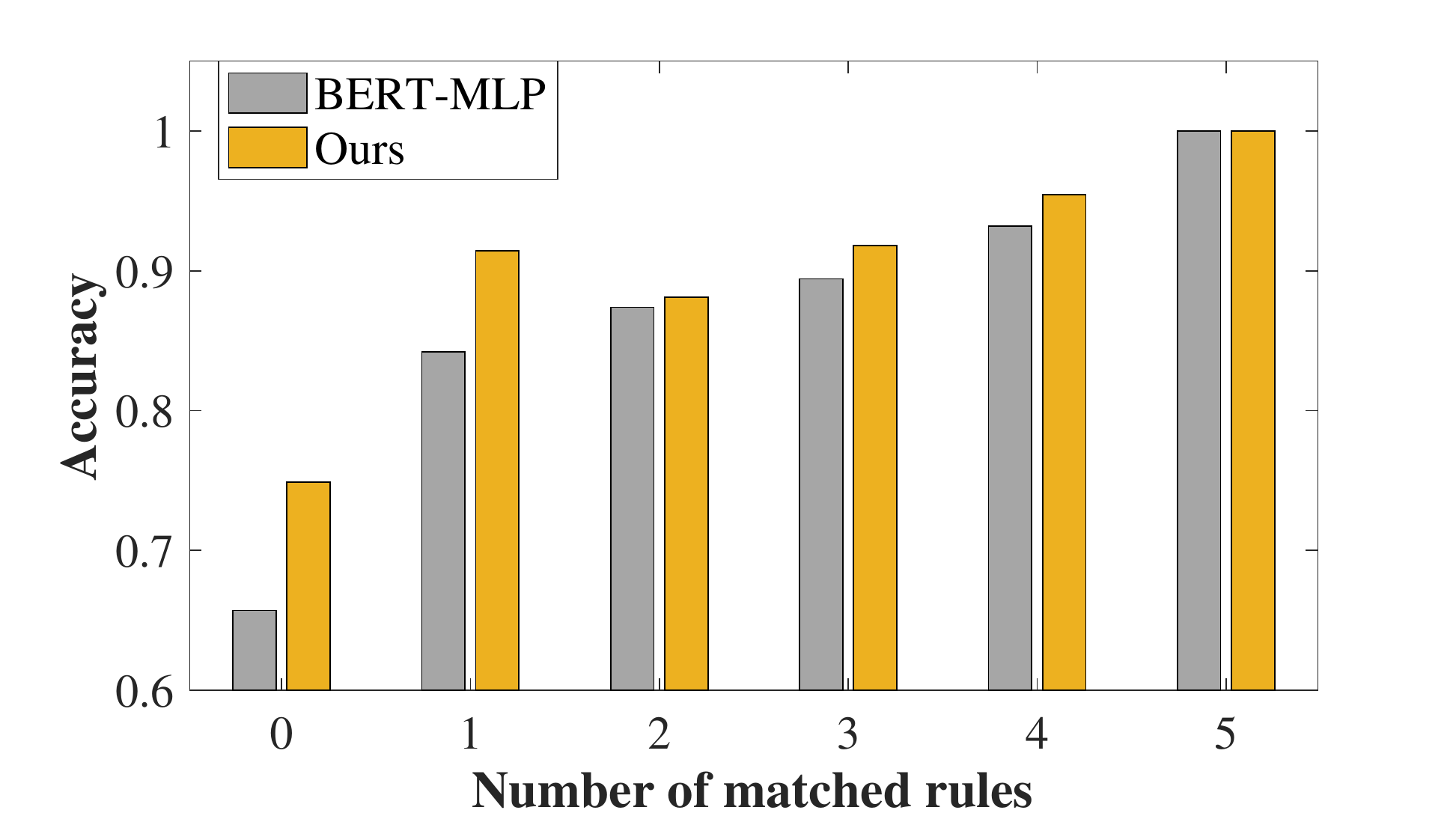}
    \caption{Accuracy on low-resource samples (matched by a small number of rules) in Youtube dataset.}
    \label{fig:change_lf_number}
\end{figure}

We proceed to study how effective our model is when dealing with the low-coverage
issue of weak supervision. To this end, we evaluate the performance of our model
for the samples covered by different numbers of rules. As shown in Figure
\ref{fig:change_lf_number}, the strongest baseline (BERT-MLP) trained with majority-voted labels
performs poorly on samples that are matched by few rules or even no rules. In
contrast, after applying our model, the performance on those less matched
samples improves significantly. This is due to the neural classifier in our
model, which predicts soft labels for unmatched samples and utilizes the
information from the multiple sources through co-training.

\subsubsection{Incorporating Clean Labels}
We also study how our model can further benefit from a small amount of labeled
data. While our model uses weak labels by default, it can easily incorporate
clean labeled data by changing the weak labels to clean ones and fix them during
training. We study the performance of our model in this setting, and compare
with the fully-supervised BERT-MLP model trained with the same amount of clean
labeled data.

\begin{table}[h]
\begin{small}
    	\begin{center}
		\begin{tabular}{ @{\hskip1.0pt} c @{\hskip1.0pt} |@{\hskip1.0pt} c @{\hskip1.0pt}|@{\hskip1.0pt} c @{\hskip1.0pt}|@{\hskip2.0pt} c @{\hskip2.0pt}|@{\hskip2.0pt} c@{\hskip2.0pt} |@{\hskip2.0pt} c@{\hskip2.0pt}|@{\hskip2.0pt} c@{\hskip2.0pt}}
    \toprule
    \textbf{Labeled} & \textbf{Method}  & \textbf{youtube} & \textbf{imdb}  & \textbf{ yelp }  & \textbf{agnews} & \textbf{spouse} \\
    \midrule
    0.5\% & Bert-MLP & 80.6 & 76.9 & 86.2 & 82.6 & 68.2 \\
    & Ours & \textbf{92.4} & \textbf{81.9} & \textbf{87.5} & \textbf{86.4} & \textbf{81.3}\\
    \hline
    2\% & Bert-MLP & 83.2 & 78.8 & 87.4& 84.7 & 72.3\\
    & Ours & \textbf{92.9} & \textbf{83.1} & \textbf{87.6} & \textbf{85.7} & \textbf{81.3} \\
    \hline
    5\% & Bert-MLP & 87.7 & 83.6 & 89.0 & 86.4 & 74.8\\
    & Ours & \textbf{93.8} & \textbf{86.1} & \textbf{90.4} & \textbf{88.2} & \textbf{82.1} \\
    \hline
    20\% & Bert-MLP & 90.8 & 86.0 & 90.3 & \textbf{89.2} & 75.6\\
    & Ours & \textbf{94.0} & \textbf{86.1}& \textbf{90.5} & \textbf{89.2} & \textbf{84.5}\\
    \hline
    50\% & Bert-MLP & 91.8 & \textbf{86.2} & \textbf{90.5} & 89.2 & 78.0\\
    & Ours & \textbf{95.4} & \textbf{86.2} & \textbf{90.5} & \textbf{89.3} & \textbf{85.9}\\
    \hline
    100\% & Bert-MLP & 94.4 & 87.2 & 91.1 & 90.7 & 79.6\\
    \bottomrule
    \end{tabular}
    \end{center}
    \caption{The classification accuracy of BERT-MLP and our model with ground truth labeled data}
    \label{tab:noise_accuracy}
\end{small}
\end{table}

As shown in Table \ref{tab:noise_accuracy}, the results of combining our
denoised labels with a small amount of clean labels are inspiring: it further
improves the performance of our model and consistently outperforms the fully
supervised BERT-MLP model. When the labeled ratio is small, the performance
improvement over the fully-supervised model is particularly large: improving the
accuracy by 6.28\% with 0.5\% clean labels and 3.84\% with 5\% clean
labels on average. When the ratio of clean labels is large, the performance
improvements becomes marginal.

The performance improvement over the fully-supervised model is relatively smaller
on yelp and agnews datasets. The reason is likely that the
text genres of yelp and agnews are similar to the text corpora used in BERT
pre-training, making the supervised model fast achieve its peak performance
with a small amount of labeled data.



\subsubsection{Ablation Study}
We perform ablation studies to evaluate the effectiveness of the three components in our model: the label denoiser, the neural classifier, and the self-training over unmatched samples.
By removing one of them, we obtain four settings:
1) Rule-only, represents w/o neural classifier and self-training;
2) Neural-only, represents w/o label denoiser and self-training;
3) Neural-self: represents w/o label denoiser;
4) Rule-Neural: represents w/o self training.
3) and 4) are supervised by the initial simple majority voted labels. Table
\ref{tab:ablation_study} shows the results.
We find that  all the three components are key to our model, because: 1) the rule-based label denoiser iteratively obtains higher-quality pseduo labels from the weak supervision sources; 2) the neural classifier extracts extra supervision signals from unlabeled data through self-training.


\begin{table}[h]\small

		\begin{center}
				\begin{tabular}{ @{\hskip2.0pt} l @{\hskip2.0pt} |@{\hskip2.0pt} c @{\hskip2.0pt}|@{\hskip2.0pt} c @{\hskip2.0pt}|@{\hskip2.0pt} c @{\hskip2.0pt}|@{\hskip2.0pt} c@{\hskip2.0pt} |@{\hskip2.0pt} c@{\hskip2.0pt}}
				\toprule
				\textbf{Method} & \textbf{youtube} & \textbf{imdb} & \textbf{  yelp  }  & \textbf{agnews} & \textbf{spouse} \\
				\midrule
				Ours & \textbf{94.9}&\textbf{82.9}& \textbf{87.5}&\textbf{85.7}& \textbf{81.3} \\
				Rule-only & 90.3 & 73.1 & 70.2 & 63.6 & 77.2\\
				Neural-only & 77.0 & 72.5 & 81.5 & 75.8 & 70.7\\
				Neural-self & 89.3 & 81.4 & 82.9 & 81.3 & 79.7\\
				Rule-Neural & 87.2 & 80.1 & 80.8 & 84.8 & 69.9\\
				\bottomrule
			\end{tabular}
		\end{center}
	\caption{Ablation Study Results.}
	\label{tab:ablation_study}
\end{table}

\subsubsection{Case Study}

We provide a example of Yelp dataset to show the denoising process of our model.

A reviewer of says ``My husband tried this place. He was pleased with his experience and he wanted to take me there for dinner. We started with calamari which was so greasy we could hardly eat it...The bright light is the service. Friendly and attentive! The staff made an awful dining experience somewhat tolerable.'' The ground-truth sentiment should be NEGATIVE. 

This review is labeled by three rules as follows: 1) keyword-mood, \texttt{pleased} $\rightarrow$ POSITIVE; 2) keyword-service, \texttt{friendly} $\rightarrow$ POSITIVE; 3) keyword-general, \texttt{awful} $\rightarrow$ NEGATIVE. The majority-voted label is thus POSITIVE, but it is wrong.  After applying our method, the learned conditional reliability scores for the three rules are 0.1074, 0.1074, 0.2482, which emphasizes rule 3) so the denoised weighted majority voted is thus NEGATIVE, and it becomes correct.

\subsubsection{Parameter Study}
\label{sec:parameter}

  The primary parameters of our model include: 1) the dimension of hidden
  layers $d_{\rm h}$ in the label denoiser and the feature-based classifier;
  2) learning rate $lr$; 3) the weight $c_1$, $c_2$, and $c_3$ of regularization term for $\ell_1$, $\ell_2$, and $\ell_3$ in \eqref{eq:loss_func}; 
  4) We fix momentum term $\alpha = 0.6$ followed the implementation of~\citet{lainete2016}.
  By default, we set $d_{\rm h} = 128$, $lr =0.02$, and $c_1 = 0.2, c_2 = 0.7, c_3 = 0.1$ as our model achieves overall good performance with these parameters. The search space of $d_{\rm h}$ is $2^{6-9}$, $lr$ is $0.01-0.1$, $c_1$ and $c_3$ are $0.1-0.9$ (note that $c_2 = 1 - c_1 - c_3$). The hyperparameter configuration for the best performance reported in Table \ref{tab:main_result_ind} is shown in the \cref{sec:hyper}.

\begin{figure}[h]
    \centering
    \includegraphics[width = 0.48\textwidth]{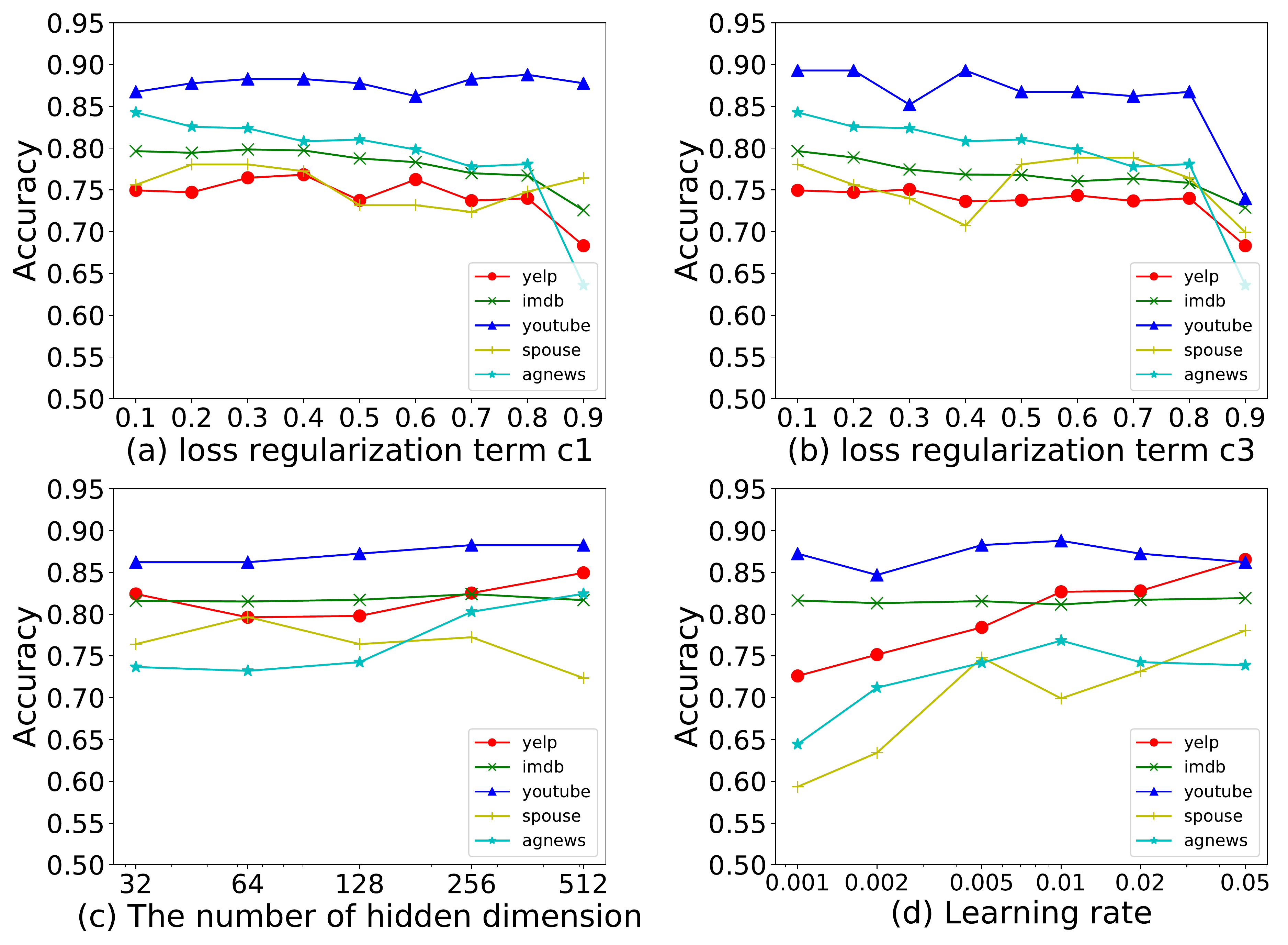}
    \caption{The prediction accuracy over different parameter settings.}
    \label{fig:parameter_c}
\end{figure}

We test the effect of one hyperparameter by fixing others to their default values.
In Figure~\ref{fig:parameter_c} (a) and (b), we find the performance is stable
except that the loss weight is too large.
For (c) and (d), except for the spouse dataset when $lr$ is too small and
$d_{\rm h}$ is too large (instability due to the dataset size is small), our
model is robust to the hyperparameters when they are in a reasonable range. We also report overall performance for all the search trails in Table \ref{tab:hyper mean var} of \cref{sec:hyper}.



\section{Related Work}
\label{sec:org34a4bb4}


\sep \textbf{Learning from Noisy Supervision.} Our work is closely related to
existing work on learning from noisy supervision. To deal with label noise,
several studies \citep{brodleyid1999, smithim2011, yangdi2018} adopt a data
cleaning approach that detects and removes mislabeled instances. This is
achieved by outlier detection \citep{brodleyid1999}, a-priori heuristics
\citep{smithim2011}, self-training \citep{bond}, or reinforcement learning \citep{yangdi2018, zhang2020seqmix}. One drawback
of this data cleaning approach is that it can discard many samples and incur
information loss.

Different from data cleaning, some works adopt a data correction approach. The
most prominent idea in this line is to estimate the noise transition matrix
among labels \citep{sukhbaatarle2014,sukhbaatartr2014,goldbergertr2016, DBLP:conf/emnlp/WangLLYL19,DBLP:journals/corr/abs-1911-00068} and
then use the transition matrices to re-label the instances or adapt the loss
functions. Specifically, \citet{DBLP:conf/emnlp/WangLLYL19} and \citet{DBLP:journals/corr/abs-1911-00068} generate label noise by flipping clean labels based on such noise transition matrices. They are thus not applicable to our weak supervision setting where no clean labels are given. Meanwhile, re-weighting strategies have been explored to
adjust the input training data. These techniques weigh training samples
according to the predictions confidence \citep{dehghaniav2017}, one-sided
noise assumption \citep{zhangle2019}, a clean set \citep{DBLP:conf/icml/RenZYU18} or the similarity of their descent
directions \citep{yangdi2018}. Recently, a few studies
\citep{veitle2017,huwe2019} have also explored designing denoising modules for
neural networks. However, our method differs from them in that: (1) our method learns \emph{conditional reliability scores} for multiple sources; and (2) these methods still require clean data for denoising, while ours does not.

\sep \textbf{Learning from Multi-Source Supervision} 
The crowdsourcing area also faces the problem of learning from multiple sources (\ie, crowd workers).
Different strategies have been proposed to integrate the annotations for the
same instance, such as estimating the confidence intervals for workers
\citep{joglekarco2015} or leveraging approval voting \citep{shahap2015}.
Compared with crowdsourcing, our problem is different in that the multiple
sources provide only feature-level noisy supervision instead of instance-level
supervision.

More related to our work are data programming methods
\citep{ratnerda2016,ratner2017snorkel,ratnertr2019} that learn from multiple
weak supervision sources. One seminal work in this line is Snorkel
\citep{ratner2017snorkel}, which treats true labels as latent variables in a
generative model and weak labels as noisy observations. The generative model is
learned to estimate the latent variables, and the denoised training data
are used to learn classifiers. Our approach differs from data programming
methods where we use a soft attention mechanism to estimate source
reliability, which is integrated into neural text classifiers to improve the performance on unmatched samples.


\sep \textbf{Self-training } Self-training is a classic
technique for learning from limited supervision \citep{yarowskyun1995}. The key
idea is to use a model's confident predictions to update the model itself
iteratively. However, one major drawback of self-training is that it is
sensitive to noise, \ie, the model can be mis-guided by its own wrong
predictions and suffer from error propagation \citep{guoon2017}.

Although self-training is a common technique in semi-supervised learning, only a few works like WeSTClass \citep{meng2018weakly} have applied it to weakly-supervised learning.
Our self-training differs from WeSTClass in two aspects: 1) it performs weighted aggregation of the predictions from multiple sources, which generates higher-quality pseudo labels and makes the model less sensitive to the error in one single source; 2) it uses temporal ensembling, which aggregates historical pseudo labels and alleviates noise propagation.

\section{Conclusion}


We have proposed a deep neural text classifier learned not from excessive
labeled data, but only unlabeled data plus weak supervisions.
Our model learns from multiple weak supervision sources using two components
that co-train each other: (1) a label denoiser that estimates source reliability
to reduce label noise on the matched samples, (2) a neural classifier that
learns distributed representations and predicts over all the samples. The two
components are integrated into a co-training framework to benefit from each
other. In our experiments, we find our model not only outperforms
state-of-the-art weakly supervised models, but also benefits supervised models
with its denoised labeled data. Our model makes it possible to train accurate
deep text classifiers using easy-to-provide rules, thus appealing in
low-resource text classification scenarios. As future work, we
are interested in denoising the weak supervision further with automatic rule
discovery, as well as extending the co-training framework to other tasks beyond
text classification.



\section*{Acknowledgments}

This work was supported in part by the National Science Foundation award III-2008334, Amazon Faculty Award, and Google Faculty Award.

\bibliographystyle{acl_natbib}

\bibliography{references}

\appendix



\clearpage

\section{Supplemental Material}

\subsection{Dataset Preparation}

We randomly split the full datasets into three parts -- a
training set, a validation set and a test set, with ratios of 80\%, 10\% and
10\%, respectively. The splitting is fixed for all the methods for fair comparisons. We use the training set to train the model, the validation set
to for optimal early stopping and hyperparameters fine-tuning, and finally
evaluate different methods on the test set.

Recall our definition of the matched corpus $\D_L$. In practice, we only regard instances covered by more than $p$ sources as ``matched'' instances, where $p\in [0, 1, 2, \dots k-1]$. Specifically, $p$ is set to $2,1,1,0,0$ for YouTube, Yelp, IMDB, AGNews, and Spouse datasets.

We obtain the pre-trained BERT embeddings from the `bert-base-uncased' model. Our pre-processed data with the BERT embeddings and weak labels are available to download at \url{https://drive.google.com/drive/u/1/folders/1MJe1BJYNPudfmpFxCeHwYqXMx53Kv4h_}. The dataset description can be found in our Github repo \url{https://github.com/weakrules/Denoise-multi-weak-sources/blob/master/README.md}.

\subsection{Model Training}
\paragraph{Computing infrastructure} Our code can be run on either CPU or GPU environment with Python 3.6 and Pytorch.

\paragraph{Running time} Our model consists of two simple MLP networks with \textbf{0.199M} trainable parameters, thus the model is very time efficient with the avearge running time \textbf{4.5 minutes}. The running time differ based on the dataset size. We test our code on the System Ubuntu 18.04.4 LTS with
\textit{CPU: Intel(R) Xeon(R) Silver 4214 CPU @ 2.20GHz} and \textit{
GPU: NVIDIA GeForce RTX 2080}. All the models are trained for a maximum of 500 epochs.
\begin{table}[h]\small
	\begin{center}
	\begin{tabular}{ @{\hskip2.0pt} c @{\hskip2.0pt} |@{\hskip2.0pt} c @{\hskip2.0pt}|@{\hskip2.0pt} c @{\hskip2.0pt}|@{\hskip2.0pt} c @{\hskip2.0pt}|@{\hskip2.0pt} c@{\hskip2.0pt} |@{\hskip2.0pt} c@{\hskip2.0pt}}
\toprule
\textbf{Dataset}  & \textbf{youtube} & \textbf{imdb} & \textbf{ yelp }  & \textbf{agnews} & \textbf{spouse} \\
\midrule
Running time (min) & 1.9 & 3.65 & 3.92 & 11.92 & 1.5  \\
\bottomrule
\end{tabular}
\end{center}
\caption{Running time for one experiment on CPU for five datasets in minutes}
\label{tab:running time}
\end{table}

\paragraph{Validation performance}
For the main results in Table \ref{tab:main_result_ind}, the corresponding validation accuracy for our model is shown in Table \ref{tab:val main}.

\begin{table}[h]\small
	\begin{center}
	\begin{tabular}{ @{\hskip2.0pt} c @{\hskip2.0pt} |@{\hskip2.0pt} c @{\hskip2.0pt}|@{\hskip2.0pt} c @{\hskip2.0pt}|@{\hskip2.0pt} c @{\hskip2.0pt}|@{\hskip2.0pt} c@{\hskip2.0pt} |@{\hskip2.0pt} c@{\hskip2.0pt}}
\toprule
\textbf{Dataset}  & \textbf{youtube} & \textbf{imdb} & \textbf{ yelp }  & \textbf{agnews} & \textbf{spouse} \\
\midrule
Validation accuracy & 87.8 & 81.8 & 88.2 & 85.6 & 79.7  \\ \hline
Test accuracy &{94.9}&{82.9}& {87.5}&{85.7}& {81.3}  \\
\bottomrule
\end{tabular}
\end{center}
\caption{validation accuracy on for five datasets of the main results in Table \ref{tab:main_result_ind}.}

\label{tab:val main}
\end{table}

\subsection{Hyperparameter Search}
\label{sec:hyper}
Since our datasets are well balanced, we use \textit{accuracy} as the criterion for optimal early stopping and hyperparameters fine-tuning. Our hyperparameter values are uniform sampled within a reasonable range with particular numbers in Table \ref{tab:range}.

\begin{table}[h]\small
	\begin{center}
	\begin{tabular}{ @{\hskip2.0pt} c @{\hskip2.0pt} |@{\hskip2.0pt} c @{\hskip2.0pt}}
\toprule
\textbf{Parameters}  & \textbf{Search Range} \\
\midrule
$d_{\rm h}$ & 32, 64, 128, 256, 512  \\
$lr$ & 0.001, 0.002, 0.005, 0.01, 0.02, 0.05 \\
$c_1$ & 0.1, 0.2, 0.3, 0.4, 0.5, 0.6, 0.7, 0.8, 0.9 \\
$c_3$ & 0.1, 0.2, 0.3, 0.4, 0.5, 0.6, 0.7, 0.8, 0.9 \\
\bottomrule
\end{tabular}
\end{center}
\caption{The hyper parameters search bounds.}
\label{tab:range}
\end{table}

Table \ref{tab:best_parameter} shows the hyper parameters used to get the best results for Table \ref{tab:main_result_ind}.

\begin{table}[h]\small
	\begin{center}
	\begin{tabular}{ @{\hskip2.0pt} c @{\hskip2.0pt} |@{\hskip2.0pt} c @{\hskip2.0pt}|@{\hskip2.0pt} c @{\hskip2.0pt}|@{\hskip2.0pt} c @{\hskip2.0pt}|@{\hskip2.0pt} c@{\hskip2.0pt} |@{\hskip2.0pt} c@{\hskip2.0pt}}
\toprule
\textbf{Parameters}  & \textbf{youtube} & \textbf{imdb} & \textbf{ yelp }  & \textbf{agnews} & \textbf{spouse} \\
\midrule
$d_{\rm h}$ & 128 & 64 & 128 & 256 & 256  \\
$lr$ & 0.02 & 0.02 & 0.02 & 0.05 & 0.02\\
$c_1$ & 0.2 & 0.2 & 0.2 & 0.1 & 0.2 \\
$c_3$ & 0.1 & 0.2 & 0.2 & 0.1 & 0.1 \\
\bottomrule
\end{tabular}
\end{center}
\caption{The hyper parameters setting for the best accuracy results of Table \ref{tab:main_result_ind}.}
\label{tab:best_parameter}
\end{table}

For the above four parameters with their range, we perform 1350 search trails. The test and validation results accuracy with mean and standard deviation for hyperparameters search experiments are in Table \ref{tab:hyper mean var}.

\begin{table}[h]\small
	\begin{center}
	\begin{tabular}{ @{\hskip2.0pt} c @{\hskip2.0pt} |@{\hskip2.0pt} c @{\hskip2.0pt}|@{\hskip2.0pt} c @{\hskip2.0pt}|@{\hskip2.0pt} c @{\hskip2.0pt}|@{\hskip2.0pt} c@{\hskip2.0pt} |@{\hskip2.0pt} c@{\hskip2.0pt}}
\toprule
\textbf{}  & \textbf{youtube} & \textbf{imdb} & \textbf{ yelp }  & \textbf{agnews} & \textbf{spouse} \\
\midrule
Val Mean  & 81.5 & 77.1 & 79.1 & 80.0 & 83.5 \\
Val Stdev & 0.019 & 0.036& 0.034 & 0.073 & 0.093\\\hline
Test Mean & 87.1 & 78.0& 81.2 & 79.8 & 79.5\\
Test Stdev & 0.021 & 0.031 & 0.042 & 0.070 & 0.118\\

\bottomrule
\end{tabular}
\end{center}
\caption{The validation and test results for the hyperparameters search trails with the mean and standard deviation.}
\label{tab:hyper mean var}

\end{table}

\subsection{Labeling Sources}
\label{sec:LFs}
We have four types of annotation rules which are Keyword Labeling Sources, Pattern-matching (Regular Expressions) Labeling Sources, Heuristic Labeling Sources, and Third-party Tools. For the first and second one, we give the uniform definitions for all the datasets.

\begin{itemize}
	\item \textit{Keyword Labeling Sources}

	Given $x$ as a document $\bm{d_i}$ in a corpus of text documents \(\D\), a keywords list $L$, and a class label \(C\) in the set of target classes \(\C\), we define keywords matching annotation process \texttt{HAS} as

    \begin{definition}[Keywords rules]

    \texttt{HAS(x, L) $\Rightarrow$ C} if $x$ matches one of the words in the list $L$.
    \end{definition}

	\item \textit{Pattern-matching Labeling Sources}

    Given $x$, a regular expression $R$, and a class label $C$, we define the pattern-matching annotation process \texttt{MATCH} as
    \begin{definition}[Pattern-matching rules]

    \texttt{MATCH(x, R) $\Rightarrow$ C} if $x$ matches the regular expression $R$.
    \end{definition}

\end{itemize}

For the remaining third and fourth types, each dataset has specific definitions. We then state all the labeling rules for each dataset from Table \ref{tab:youtube} to Table \ref{tab:spouse}.

\subsubsection{Statistics of Labeling Sources}

\begin{figure}[htb!]
    \centering
    \includegraphics[width = 0.43\textwidth]{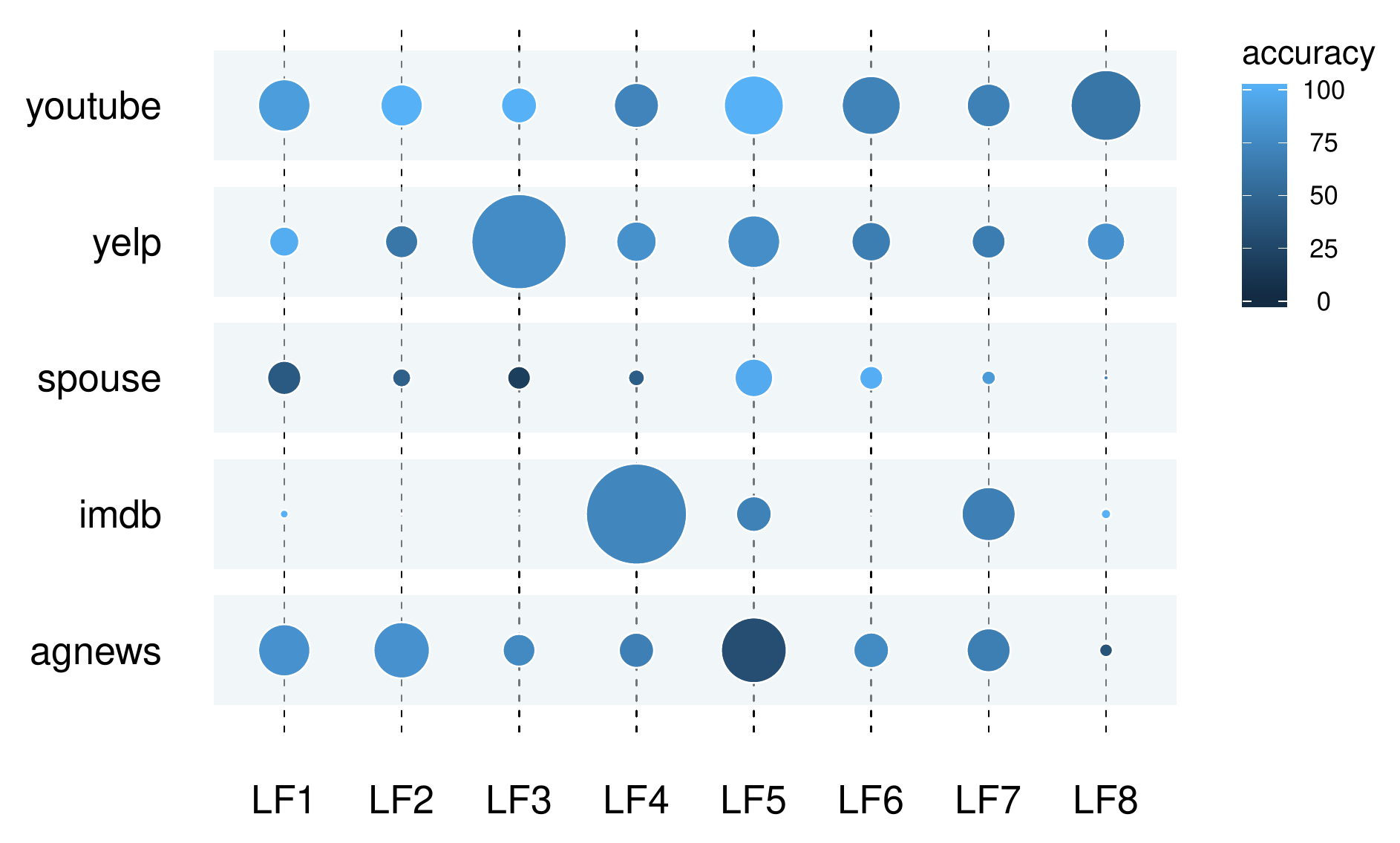}
    \caption{The coverage and accuracy of our used labeling functions on five datasets. Larger circle denotes higher coverage and lighter color denotes higher accuracy.}
    \label{fig:lf_analysis}
\end{figure}

We show the accuracy and coverage of each rule in the Fig \ref{fig:lf_analysis}, where the shape represents the coverage and the color depth represents the accuracy of the rule-induced labeled data. The average accuracy of these rules is 67.5\%, and the average coverage is 23.3\%.


We also show one example of Yelp dataset with the detail statistics for each labeling source, and the rule descriptions are in Table \ref{tab:yelp}.

\begin{table}[htb!]\small
    \centering
    \begin{tabular}{c|c|c}
    \toprule
         \textbf{Labeling source} & \textbf{Coverage} & \textbf{Emp. Accu}\\ \midrule
  textblob             & 6.80               & 97.06              \\
  keyword\_recommand   & 8.40               & 59.52              \\
  keyword\_general     & 75.20              & 74.20              \\
  keyword\_mood        & 12.80              & 78.12              \\
  keyword\_service     & 33.30              & 75.68              \\
 keyword\_price       & 23.30              & 63.93              \\
  keyword\_environment & 8.80               & 63.64             \\
  keyword\_food        & 11.40              & 78.95              \\ \bottomrule

    \end{tabular}
    \caption{The labeling rules statictics for Yelp dataset. Both Coverage and Emp. Accu (number of corrected samples / rule-matched samples) are in \%.}
    \label{tab:my_label}
\end{table}

\subsubsection{Rules Description}
We show some examples of labeling rules here, and the full description of rules and their corresponding weak labels are in our Github repo \url{https://github.com/weakrules/Denoise-multi-weak-sources/tree/master/rules-noisy-labels}.

\paragraph{Youtube} 
We use the same labeling functions as \citep{ratner2017snorkel}, and we show the rules with an example in Table \ref{tab:youtube}.

\begin{table*}[htb!]
    \centering
    \begin{tabular}
    {p{190pt} | p{230pt} }
    \toprule
        \textbf{Rule} & \textbf{Example} \\
        \midrule
        \texttt{HAS(x,[my]) $\Rightarrow$ SPAM}
        & Plizz withing my channel  \\ \hline
        \texttt{HAS(x, [subscribe]) $\Rightarrow$ SPAM} & Subscribe to me and I'll subscribe back!!\\ \hline
        \texttt{HAS(x, [http]) $\Rightarrow$ SPAM} & 
        please like : http://www.bubblews.com/news/9277547-peace-and-brotherhood
        \\\hline
        \texttt{HAS(x, [please, plz]) $\Rightarrow$ SPAM} & 
        Please help me go here http://www.gofundme.com/littlebrother \\\hline
        \texttt{HAS(x, [song]) $\Rightarrow$ HAM} & 
        This song is great there are 2,127,315,950 views wow \\\hline
        \texttt{MATCH(x, check.*out ) $\Rightarrow$ SPAM} & Please check out my vidios\\\hline
        We define \texttt{LENGTH(x)} as the number of words in $x$.\\
        \texttt{LENGTH(x) < 5 $\Rightarrow$ HAM} & 
        2 BILLION!! \\\hline
        We define $x.ents$ as the tokens of $x$, and $x.ent.label$ as its label.\\
        $LENGTH(x) < 20 \quad AND \quad any([ent.label == PERSON \ for \ ent\  in\  x.ents] \Rightarrow HAM$ & 
        Katy Perry is garbage. Rihanna is the best singer in the world. \\\hline
        We define \texttt{POLARITY(x)} as the sentiment subjectivity score obtained from the TextBlob tool, a pretrained sentiment analyzer.\\
        $POLARITY(x) > 0.9 \Rightarrow HAM$ &
    Discover a beautiful song of A young Moroccan http://www.linkbucks.com/AcN2g \\
    \bottomrule
    \end{tabular}
    \caption{Youtube labeling sources examples}
    \label{tab:youtube}
\end{table*}

\paragraph{IMDB}

The rules are straightforward so we show the rules without the sentence examples in Table \ref{tab:imdb}.

\begin{table*}[htb!]
    \centering
    \begin{tabular}
    {p{430pt} }
    \toprule
        \textbf{Rule} \\
        \midrule
    \texttt{[masterpiece, outstanding, perfect, great, good, nice, best, excellent, worthy, awesome, enjoy, positive, pleasant, wonderful, amazing, superb, fantastic, marvellous, fabulous] $\Rightarrow$ POS}
     \\ \hline
    \texttt{[bad, worst, horrible, awful, terrible, crap, shit, garbage, rubbish, waste] $\Rightarrow$ NEG} \\ \hline
    \texttt{[beautiful, handsome, talented]$\Rightarrow$ POS}
     \\ \hline
    \texttt{ [fast forward, n t finish] $\Rightarrow$ NEG}  \\ \hline
    \texttt{[well written, absorbing, attractive, innovative, instructive, interesting, touching, moving]$\Rightarrow$ POS}
     \\ \hline
    \texttt{[to sleep, fell asleep, boring, dull, plain]$\Rightarrow$ NEG}  \\ \hline
    \texttt{[ than this,  than the film,  than the movie]$\Rightarrow$ NEG}  \\ \hline
    \texttt{MATCH(x, *PRE*EXP* ) $\Rightarrow$ POS} \\
    \texttt{PRE = [will,  ll , would ,  d , can t wait to ]}\\
    \texttt{EXP = [next time, again, rewatch, anymore,  rewind]}
     \\ \hline
    \texttt{MATCH(x, *PRE*EXP* ) $\Rightarrow$ POS} \\
    \texttt{PRE = [highly, do, would, definitely, certainly, strongly, i, we]}\\
    
    \texttt{EXP = [recommend,  nominate]}
     \\ \hline
    \texttt{MATCH(x, *PRE*EXP* ) $\Rightarrow$ POS} \\
    \texttt{PRE = [high, timeless, priceless, has, great, real, instructive]}\\
    \texttt{EXP = [value, quality, meaning, significance]} \\ 

    \bottomrule
    \end{tabular}
    \caption{IMDB labeling sources examples}
    \label{tab:imdb}
\end{table*}

\paragraph{Yelp}

The rules are straightforward so we show the rules without the sentence examples in Table \ref{tab:yelp}. We provide labeling rules in eight views.

\begin{table*}[htb!]
    \centering
    \begin{tabular}
    { p{50pt} | p{350pt} }
    \toprule
        \textbf{View}  & \textbf{Rule} \\
        \midrule
        General & \texttt{[outstanding, perfect, great, good, nice, best, excellent, worthy, awesome, enjoy, positive, pleasant,wonderful, amazing] $\Rightarrow$ POS}
        \\ \hline
        General & \texttt{[bad, worst, horrible, awful, terrible, nasty, shit, distasteful,dreadful, negative]$\Rightarrow$ NEG} \\ \hline
        
        Mood & \texttt{[happy, pleased, delighted,contented, glad, thankful, satisfied] $\Rightarrow$ POS}
        \\ \hline
        Mood & \texttt{[sad, annoy, disappointed,frustrated, upset, irritated, harassed, angry, pissed]$\Rightarrow$ NEG}  \\ \hline
        Service & \texttt{[friendly, patient, considerate, enthusiastic, attentive, thoughtful, kind, caring, helpful, polite, efficient, prompt] $\Rightarrow$ POS}
        \\ \hline
        Service & \texttt{[slow, offended, rude, indifferent, arrogant]$\Rightarrow$ NEG} \\ \hline
        Price & \texttt{[cheap, reasonable, inexpensive, economical] $\Rightarrow$ POS}
        \\ \hline
        Price& \texttt{[overpriced, expensive, costly, high-priced]$\Rightarrow$ NEG}  \\ \hline
                Environment & \texttt{[clean, neat, quiet, comfortable, convenien, tidy, orderly, cosy, homely] $\Rightarrow$ POS}
        \\ \hline
        Environment & \texttt{[noisy, mess, chaos, dirty, foul]$\Rightarrow$ NEG}  \\ \hline
                Food & \texttt{[tasty, yummy, delicious,appetizing, good-tasting, delectable, savoury, luscious, palatable] $\Rightarrow$ POS}
        \\ \hline
        Food & \texttt{[disgusting, gross, insipid]$\Rightarrow$ NEG} \\ \hline
              &  \texttt{[recommend] $\Rightarrow$ POS}
        \\ \hline
        Third-party & $POLARITY(x) > 0.5 \Rightarrow$ POS \\ 
         Tools & $POLARITY(x) > 0.5 \Rightarrow$ NEG \\
    \bottomrule
    \end{tabular}
    \caption{Yelp labeling sources examples}
    \label{tab:yelp}
\end{table*}

\paragraph{AGnews}
The rules are straightforward so we show the rules without the sentence examples in Table \ref{tab:agnews}.

\begin{table*}[htb!]
    \centering
    \begin{tabular}
    { p{420pt} }
    \toprule
        \textbf{Rule} \\
        \midrule
        \texttt{[ war , prime minister, president, commander, minister,  annan, military, militant, kill, operator] $\Rightarrow$ POLITICS}
         \\ \hline

        \texttt{[baseball, basketball, soccer, football, boxing,  swimming, world cup, nba,olympics,final, fifa] $\Rightarrow$ SPORTS}
         \\ \hline
        
        \texttt{[delta, cola, toyota, costco, gucci, citibank, airlines] $\Rightarrow$ BUSINESS}
         \\ \hline

        \texttt{[technology, engineering, science, research, cpu, windows, unix, system, computing,  compute] $\Rightarrow$ TECHNOLOGY }
         \\
    \bottomrule
    \end{tabular}
    \caption{AGnews labeling sources examples}
    \label{tab:agnews}
\end{table*}

\paragraph{Spouse}

We use the same rule as \citep{ratner2017snorkel} and we show the definition as well as examples in Table \ref{tab:spouse}.

\begin{table*}[htb!]
    \centering
    \begin{tabular}
    {p{260pt} | p{160pt} }
    \toprule
        \textbf{Rule} & \textbf{Example} \\
        \midrule
        \texttt{[father, mother, sister, brother, son, daughter, grandfather, grandmother, uncle, aunt, cousin] $\Rightarrow$ NEG}
        & His 'exaggerated' sob stories allegedly include claiming he had cancer, and that his son had made a suicide attempt.  \\ \hline

        \texttt{[boyfriend, girlfriend, boss, employee, secretary, co-worker] $\Rightarrow$ NEG}
        & Dawn Airey’s departure as European boss of Yahoo after just two years will bring a smile to the face of Armando Iannucci. \\ \hline
        \texttt{MATCH(x, *PERSON1*LIST*PERSON2* $\Rightarrow$ POS}
        \texttt{LIST = [spouse, wife, husband, ex-wife, ex-husband]}
        & On their wedding day, last week sundayGhanaian actress Rose Mensah, popularly known as Kyeiwaa, has divorced her husband Daniel Osei, less than four days after the glamorous event.\\ \hline
        We define \texttt{LASTNAME(x)} as the last name of x.
        \texttt{LASTNAME(person1) == LASTNAME(person2) $\Rightarrow$ POS}
        & Karen Bruk and Steven Bruk, Mrs. Bruk's spouse, exercise shared investment power over the Shares of the Company held by Karen Bruk and KMB.\\
    \bottomrule
    \end{tabular}
    \caption{Spouse labeling sources examples}
    \label{tab:spouse}
\end{table*}

\end{document}